%% file: acl_latex.tex
\pdfoutput=1

\documentclass[11pt]{article}

\usepackage[final]{acl}

\usepackage{times}
\usepackage{latexsym}
\usepackage{amsmath}
\usepackage{listings}
\usepackage[T1]{fontenc}

\usepackage[utf8]{inputenc}

\usepackage{microtype}

\usepackage{inconsolata}

\usepackage{graphicx}

\usepackage{booktabs}
\usepackage{todonotes}

\title{Overview of the 2024 ALTA Shared Task: Detect Automatic AI-Generated Sentences for Human-AI Hybrid
Articles}

\author{Diego Moll\'a \and Qiongkai Xu\\
  Macquarie University\\
  Sydney, Australia\\
  \texttt{diego.molla-aliod@mq.edu.au} \\
  \texttt{qiongkai.xu@mq.edu.au}\\\And
  Zijie Zeng \and 
Zhuang Li\\
  Monash University \\
  Melbourne, Australia \\
  \texttt{zhuang.li1@monash.edu}\\
  \texttt{zijie.zeng@monash.edu}\\
}

\begin{document}
\maketitle
\begin{abstract}
The ALTA shared tasks have been running annually since 2010. In 2024, the purpose of the task is to detect machine-generated text in a hybrid setting where the text may contain portions of human text and portions machine-generated. In this paper, we present the task, the evaluation criteria, and the results of the systems participating in the shared task.

\end{abstract}

\section{Introduction}

\input{0_introduction}

\section{Related Work}

\input{1_related}

\section{Data Description}

\input{2_data}

\section{Baselines}
\input{3_baselines}

\section{Evaluation Framework}

\input{4_evaluation}

\section{Participating Systems and Results}

\input{5_participant}

\section{Conclusions}

This paper described a shared task for sentence-level detection of GPT-3.5-turbo-generated content within hybrid texts. By moving beyond traditional corpus-level detection to sentence-level analysis, this task addresses the practical challenges of identifying AI-generated sentences in collaborative writing scenarios. The multi-domain training approach, combined with a focused evaluation of news articles, provides a rigorous framework for developing and evaluating fine-grained detection methods. Through this shared task, we aim to establish benchmarks for sentence-level AI content detection and advance our understanding of the distinctive characteristics of human-AI collaborative writing.

\bibliography{custom}




\end{document}

%% file: 0_introduction.tex
The advent of large language models (LLMs) has revolutionized artificial intelligence (AI), leading to a significant surge in AI-generated text and the rise of human-AI collaborative writing. While this collaboration offers exciting opportunities, it also introduces challenges --- particularly in distinguishing between human-authored and AI-generated content within a single document. Although AI refers to various technologies, our focus in this shared task is specifically on the text generated by LLMs. Detecting such content has become essential not only as a deterrent against misuse but also as a safeguard, particularly in news reporting, journalism, and academic writing.

Previous efforts, such as the 2023 ALTA shared task~\cite{molla2023overview}, focused on corpus-level detection of AI-generated text, assuming that entire documents are either human-written or AI-generated. However, with the rise of human-AI collaborative writing, it is increasingly common for a single document to contain a mix of sentences authored by human and AI. Our proposed task addresses this realistic scenario by automatically identifying AI-generated sentences within hybrid articles.

Detecting AI-generated content at the sentence level is crucial for  analyzing hybrid texts, which are becoming more prevalent in fields like news reporting, content marketing, and academic writing~\cite{ma2023abstract}. Identifying AI-generated content at a finer granularity introduces a more nuanced challenge than distinguishing entirely AI-generated documents from those solely by human writers.

To tackle this challenge, our study leverages a newly available public dataset from~\citet{zeng2024towards} and a private test set we collected for this shared task, both of which contain diverse and realistic hybrid articles. These datasets offer ideal benchmarks for exploring AI-generated text detection, as they include a mixture of human-written and AI-generated sentences across a range of topics within two key domains: academic writing and news reporting.

By examining the accuracy of identifying AI-generated sentences within texts that combine human and AI-authored content, we aim to develop more sophisticated and effective detection methods for collaborative writing scenarios. This work complements existing corpus-level detection efforts by offering a more comprehensive approach to understanding and identifying AI-generated content at different scales and contexts. The insights gained from this shared task will be valuable not only for preserving integrity in written communication but also for promoting transparency and responsibility in AI-assisted content creation.

The website of the 2024 ALTA shared task is \url{https://www.alta.asn.au/events/sharedtask2024/}.

%% file: 1_related.tex
Recent advances in LLMs have created unprecedented challenges for content authenticity. Following the comprehensive related work presented by \citet{zeng2024towardshybrid}, we examine how the ability of AI to generate human-like text raises significant concerns across multiple scenarios --- from education and journalism to scientific research~\cite{ma2023abstract} and social media. While these technologies offer tremendous benefits, they also present risks of academic dishonesty \cite{mitchell2023detectgpt} and the potential spread of misinformation.
Current detection approaches predominantly employ binary classification at the document level \cite{Koike-OUTFOX-2024,hu2024bad,he2023DeBERTav,mitchell2023detectgpt,pagnoni2022,synscipass,li2024scar}. These methods assume the content is either entirely AI-generated or entirely human-written, an assumption that fails to reflect real-world usage patterns. As noted in emerging research \cite{dugan2023real}, modern content creation often involves human-AI collaboration, requiring more fine-grained detection approaches.
A promising direction in hybrid text analysis has emerged, focusing on the identification of mixed authorship within documents. This approach draws inspiration from classical text segmentation techniques while addressing the unique challenges of AI text detection \cite{text-seg-2023-lessons,xia2023sequence}. Recent work has explored both boundary detection methods \cite{zeng2024towards,lukasik2020text,yu2023improving,xing2020improving,li2022human,somasundaran2020two,koshorek2018text} and more sophisticated approaches that integrate boundary identification with content classification \cite{bai2023SegFormer,lo2021transformer,gong2022tipster,tepper2012statistical,zeng2024towardshybrid,seqxgpt_2023}.

%% file: 2_data.tex
For this shared task, we constructed a dataset comprising hybrid articles with mixed human-written and GPT-3.5-turbo-generated\footnote{\url{https://platform.openai.com/docs/models/gpt-3-5-turbo}} content to facilitate the evaluation of AI-generated sentence detection methods.

\paragraph{Data Production.} The training data was primarily sourced from the publicly available dataset curated by \citet{zeng2024towards}, created via systematically replacing selected sentences in human-written articles with GPT-3.5-turbo-generated alternatives. For each sentence replacement, GPT-3.5-turbo was prompted to generate a contextually appropriate substitute that preserved the coherence and style of the original article.

Additionally, we expanded the dataset by generating hybrid articles from human-written news content sourced from the \textsc{CC-News} dataset~\cite{Hamborg2017}. We randomly selected 3,000 articles with token lengths between 100 and 300 and tokenized them using the NLTK tokenizer\footnote{\url{https://www.nltk.org/api/nltk.tokenize.html}}. Following the methodology outlined by \citet{zeng2024towards}, we processed these articles by replacing selected sentences with GPT-3.5-turbo-generated content. For more details on the prompt format used, please refer to \citet{zeng2024towards}.

\paragraph{Content Structure.} Each hybrid news article includes a mix of human-written and GPT-3.5-turbo-generated sentences, with sentence-level authorship labels. We employed four distinct construction patterns to organize the human and machine-generated sentences, aligning with the methods in \citet{zeng2024towards}:

\begin{itemize}
    \item \texttt{h-m}: Human-written sentences followed by machine-generated sentences.
    \item \texttt{m-h}: Machine-generated sentences followed by human-written sentences.
    \item \texttt{h-m-h}: Human-written sentences, followed by machine-generated sentences, and then human-written sentences.
    \item \texttt{m-h-m}: Machine-generated sentences, followed by human-written sentences, and then machine-generated sentences.
\end{itemize}

\paragraph{Domain Focus.} While the training data includes both academic and news domains, the evaluation exclusively targets sentence-level predictions in the news domain.

Table~\ref{tab:datastats} presents the statistics of the training and test datasets.

\begin{table*}[h]
    \centering
    \begin{tabular}{ccrrr}
         \textbf{Dataset} & \textbf{Domain} & \textbf{Documents} & \multicolumn{2}{c}{\textbf{Sentences}}  \\
         &&& \textbf{Human} & \textbf{Machine}\\
         \midrule
         Train & Academic & 14,576 & 67,647 & 132,002 \\
         Train & News & 1,500 & 4,574 & 8,571\\
         \midrule
         Phase 1 Test & News & 500 & 1,624 & 2,640\\
         Phase 2 Test & News & 1,000 & 3,310 & 5,342\\
    \end{tabular}
    \caption{Statistics of the shared task datasets}
    \label{tab:datastats}
\end{table*}

%% file: 3_baselines.tex
To establish baseline performance metrics for the task, we have implemented three approaches for AI-generated sentence detection:

\begin{itemize}
    \item \textbf{Context-Aware BERT Classifier}: A fine-tuned BERT~\cite{kenton2019bert} model that incorporates contextual information by processing three-sentence windows (the target sentence and one sentence before and after). These contextual embeddings are passed through a feed-forward neural network with a binary classification head for authorship prediction.
    
    \item \textbf{TF-IDF Logistic Regression Classifier}: A logistic regression model trained on TF-IDF vectors computed from individual sentences. The model processes each sentence independently, using these statistical features to learn discriminative patterns between human-written and AI-generated text. This baseline has been made available to the shared task participants.\footnote{\url{https://github.com/altasharedtasks/ALTA_2024_demo}}
    
    \item \textbf{Random Guess Classifier}: A naive approach that assigns authorship labels randomly, providing a lower bound for performance evaluation.
\end{itemize}


%% file: 4_evaluation.tex
\subsection{Evaluation Setup}

The evaluation was hosted as a CodaLab competition\footnote{\url{https://codalab.lisn.upsaclay.fr/competitions/19633}} with three phases.

\begin{itemize}
    \item In phase 1 (``Development''), labelled training data was made available, together with a labelled test set to test the participant systems. The CodaLab page allowed each participant to submit up to 100 system runs based on the test set of phase 1. The evaluation results of this phase appeared in a leaderboard but were not used for the final ranking.
    \item In phase 2 (``Test''), a new unlabelled test set was made available. Each team could make up to 3 submissions, the evaluation results of which were used for the final ranking.
    \item Phase 3 (``Unofficial submissions'') was open after the end of phase 2, where participating systems can make up to 999 submissions of the output of the test set of phase 2 for final analysis. The evaluation results of phase 3 were not used for the final ranking. Phase 3 is open indefinitely, and new teams are encouraged to participate and compare their systems against the published results.
\end{itemize}

The labels of the test set used in phases 2 and 3 are not publicly available.

\subsection{Evaluation Metrics}
Participants are tasked with identifying the authorship of each sentence in a hybrid article $A$ consisting of $n$ sentences $\{s_1, s_2, \ldots, s_n\}$. Each sentence is either human-written or AI-generated. Formally, we define a function $f$ that maps the hybrid article $A$ to a sequence of predicted labels $\hat{L}$:
\begin{equation}
    f(A) \rightarrow \hat{L}, \quad \text{where} \quad \hat{L} = \{\hat{l}_1, \hat{l}_2, \ldots, \hat{l}_n\}
\end{equation}
Each label $\hat{l}_i$ indicates the predicted authorship of the corresponding sentence $s_i$, being either human-written (H) or AI-generated (A).

The performance is primarily evaluated using Cohen's Kappa score, with accuracy serving as a supplementary metric.

\paragraph{Cohen's Kappa Score.} This robust statistic, which determines the final system rankings, measures inter-rater agreement while accounting for chance agreement:
\begin{equation}
    \kappa = \frac{p_o - p_e}{1 - p_e}
\end{equation}
where $p_o$ is the observed agreement (accuracy), and $p_e$ is the expected agreement by chance. The Kappa score effectively handles imbalanced datasets where one class may dominate, making it particularly suitable for evaluating detection performance across varying distributions of human-written and AI-generated content.

\paragraph{Accuracy.} As a supplementary metric, we also report the proportion of correctly classified sentences across all test articles.

The evaluation metrics have been implemented using scikit-learn functions \verb+cohen_kappa_score+ and \verb+accuracy_score+.


%% file: 5_participant.tex
As in previous years, there were two categories of participating teams:
\begin{itemize}
    \item \textbf{Student}: All team members must be university students. No participating members can be full-time employees or have completed a PhD in a relevant field. The only exception is student supervisors.
    \item \textbf{Open}: Any other teams fall into the open category.
\end{itemize}

\begin{table*}[t]
    \centering
    \begin{tabular}{llrr}
      Team & Category & Kappa & Accuracy \\
      \midrule
       Dima & Student & 0.9416 & 0.9724\\
       SamNLP & Student & 0.9245 & 0.9642\\
       Adventure Seeker & Student & 0.8183 & 0.9163\\
       ADSN & Open & 0.6955 & 0.8548\\
    \end{tabular}
    \caption{Results of participating systems on the phase 2 evaluation set.}
    \label{tab:results}
\end{table*}

\begin{table}[t]
    \centering
    \begin{tabular}{lrr}
      Method & Kappa & Accuracy \\
      \midrule
       Context-Aware BERT & 0.8461 & 0.9294\\
       Logistic Regression & 0.5674 & 0.7973\\
       Random Guess & 0.0012 & 0.4973\\
    \end{tabular}
    \caption{Results of baseline systems on the phase 2 evaluation set.}
    \label{tab:baselines}
\end{table}

A total of 4 teams made submissions in the test phase, and the results are shown in Table~\ref{tab:results}. 
The Kappa score was used for the final ranking, while the Accuracy score is provided to facilitate comparisons with previous and future work. As shown in Table~\ref{tab:baselines}, all participating teams outperformed the logistic regression and random baselines, while two teams achieved better results than the BERT baseline.

The difference between the top team and second best is statistically different\footnote{Tests of statistical significance were based on NcNemar test on the system outputs, using the tool provided by \citet{dror-etal-2018-hitchhikers}.}, so the winning team is ``null-error''.

A brief description of the participating systems who provided their information follows.

\paragraph{Team Dima} \cite{dima-2024} used a 4-bit quantized LlaMA 3.1-8B-Instruct fine-tuned on domain-specific data. They also tested their system's ability to handle automatic rewrites.

\paragraph{Team ADSN} \cite{adsn-2024} used an ensemble of lightweight classification methods inspired on traditional authorship attribution approaches.